\documentclass[letterpaper, 10 pt, conference]{ieeeconf}  

\usepackage{algorithm}
\usepackage{algpseudocode}
\usepackage{balance}
\usepackage{epstopdf}

\usepackage{graphicx}
\graphicspath{ {./images/} }
\usepackage{subfig}

\usepackage{amsmath,amssymb,amsfonts}
\usepackage{xcolor}
\usepackage{bm}

\newcommand{\R}{\mathbb{R}} 

\usepackage[colorinlistoftodos]{todonotes}

\IEEEoverridecommandlockouts                              
\title{\LARGE \bf
Humanoid Robot Co-Design: Coupling Hardware Design with Gait Generation via Hybrid Zero Dynamics
}

\author{Adrian B. Ghansah$^{1}$, Jeeseop Kim$^{2}$, Maegan Tucker$^{2}$, and Aaron D. Ames$^{1,2}$
\thanks{The work is supported by National Science Foundation (CPS Award \#1932091), the Technology Innovation Institute, and the Zeitlin Family Fund}%
\thanks{$^{1}$ Authors are with the Department of Control and Dynamical Systems, California Institute of Technology,
Pasadena, CA 91125, USA, {\tt \small \{aghansah, ames\}@caltech.edu}}
\thanks{$^{2}$ Authors are with the Department
of Mechanical and Civil Engineering, California Institute of Technology,
Pasadena, CA 91125, USA, {\tt \small \{jeeseop, mtucker, ames\}@caltech.edu}}%
}

\begin{document}

\maketitle
\thispagestyle{empty}
\pagestyle{empty}

\begin{abstract}
Selecting robot design parameters can be challenging since these parameters are often coupled with the performance of the controller and, therefore, the resulting capabilities of the robot. This leads to a time-consuming and often expensive process whereby one iterates between designing the robot and manually evaluating its capabilities. This is particularly challenging for bipedal robots, where it can be difficult to evaluate the behavior of the system due to the underlying nonlinear and hybrid dynamics. Thus, in an effort to streamline the design process of bipedal robots, and maximize their performance, this paper presents a systematic framework for the co-design of humanoid robots and their associated walking gaits. 
To this end, we leverage the framework of hybrid zero dynamic (HZD) gait generation, which gives a formal approach to the generation of dynamic walking gaits.  The key novelty of this paper is to consider both virtual constraints associated with the actuators of the robot, coupled with \emph{design} virtual constraints that encode the associated parameters of the robot to be designed.  
These virtual constraints are combined in an HZD optimization problem which simultaneously determines the design parameters while finding a stable walking gait that minimizes a given cost function. 
The proposed approach is demonstrated through the design of a novel humanoid robot, ADAM, wherein its thigh and shin are co-designed so as to yield energy efficient bipedal locomotion. 
\end{abstract}

\section{INTRODUCTION}
Humanoid robots have been of great interest in the field of robotics for several decades. Because our world is built for humans, it would be advantageous to have robots with morphology similar to ours. With the emergence of cheap actuators capable of handling dynamic walking \cite{katz2018low}, recent work has been aimed at developing dynamic humanoid robot platforms \cite{chignoli2021humanoid}, \cite{sim2022tello}. While these platforms are promising, they are challenging to develop because of their complexity, both in terms of the mechanical design as well as controller synthesis. The mechanical intricacies lie in their need to be lightweight yet robust to impacts. Furthermore, it is desirable to have large ranges of motion in order to facilitate as many different behaviors as possible. On the other hand, developing controllers for humanoids is challenging due to the inherent instability and the nonlinear and hybrid nature of legged locomotion. Due to these challenges, it is common to approach the design and control problems separately, which often leads to an iterative and time-consuming process. 

\begin{figure}
    \centering
    {{\includegraphics[width = 0.95\linewidth]{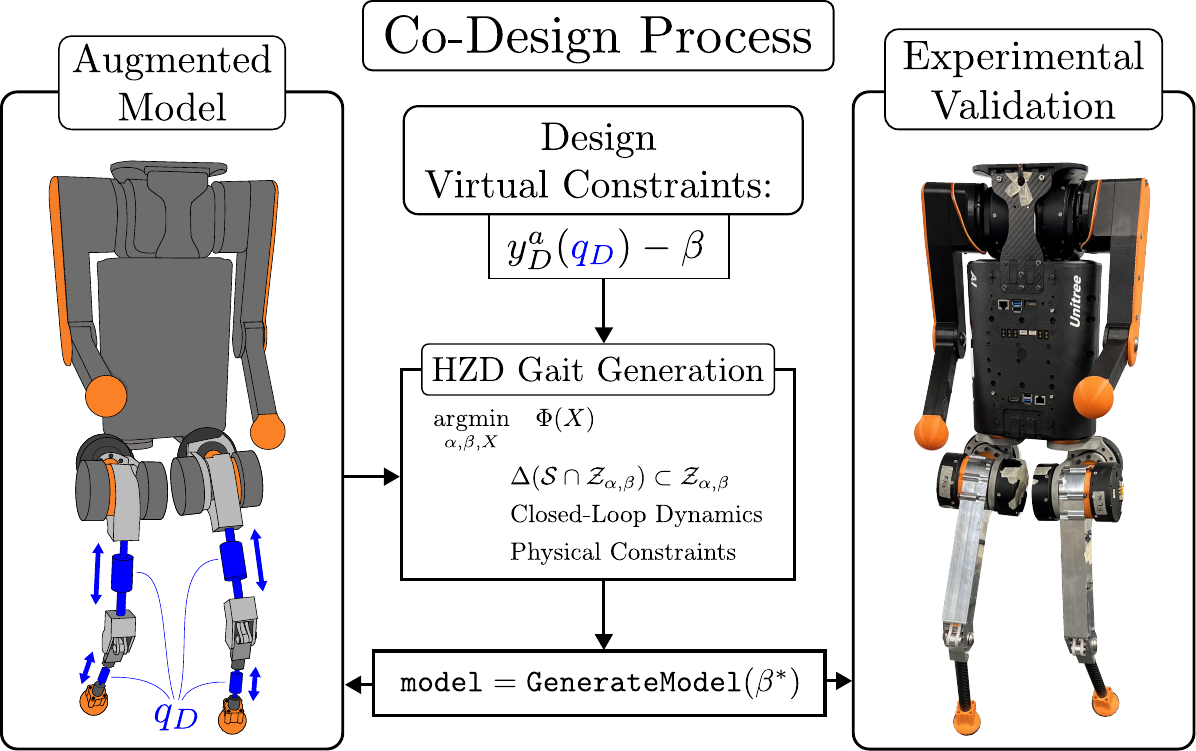}}}
    \caption{The proposed HZD-Co-Design framework is demonstrated towards systematically selecting design parameters (i.e. the leg dimensions) for the humanoid ADAM. 
    }
    \label{fig:humanoid_front_page}
    \vspace{-6mm}
\end{figure}

\subsection{Legged Robot Co-Design: Background and Related Work}

One common approach to designing legged robots is to take inspiration from nature. Various types of biologically inspired robots have been developed over the years, including cheetahs \cite{ananthanarayanan2012towards}, chimpanzees \cite{kuehn2014distributed}, and kangaroos \cite{graichen2015control}. While biologically inspired designs can aid greatly in the overall development of a robot, it is not always straightforward or even feasible, as the form of actuation and tasks to complete of the robotic system may be very different from their counterparts in nature. For these reasons, researchers have investigated different ways of letting the design process be guided by the control objectives and tasks that the robot is intended to complete, a method often referred to as concurrent design (co-design) \cite{li2001design}. 

Several interesting model-based co-design methodologies have been developed. In \cite{ha2016task} the authors propose a  framework where the goal is to find a relationship between a robot's dimensions and the tasks it intends to achieve. This is done as a bi-level optimization problem where the first step is to generate motion plans for a simplified model containing the center of mass and the feet locations, and then in the second step, the link lengths and the full-body motions are optimized to achieve the desired task. In \cite{chadwick2020vitruvio} the Vitruvio toolbox is presented, which uses a single rigid body dynamics (SRBD) model to generate motion plans for the robot to determine predefined forces and motions, and a genetic algorithm is used to find the optimal leg parameters. 

While the aforementioned design frameworks aid the design process, they rely on reduced-order models which limits their application to systems where the massless limb assumptions hold reasonably well, such as most quadrupedal robots \cite{di2018dynamic}. For humanoid robots, the design often necessitates heavier legs since the entire weight of the robot needs to be supported by the two legs, especially when carrying heavy payloads. 
In this context, there have been examples of co-design that have leveraged the full-order system dynamics. 
For example, in \cite{ambrose2021towards}, a full-order model was used to find optimal spring variables for a hopping robot concurrently with generating the hopping motion. This was done by modeling the interactions between the robot's base and the springs as an interconnected system \cite{antonelli2013interconnected} and then generating hopping for the co-designed system by solving the resulting optimization problem.


Lastly, it is important to note the existence of model-free co-design methods. An example of this was done by \cite{belmonte2022meta}, where reinforcement learning \cite{sutton2018reinforcement} was used. The work builds on \cite{won2019learning} and does not rely on predefined control schemes and rather attempts to provide an optimal design using an end-to-end solution. An advantage of the method is that it seems to perform better when dealing with unexpected changes in the environment and the optimized designs are less likely to be overfitted to certain motions. However, because these approaches do not take advantage of model information, they can be computationally expensive with long run-times.

\subsection{Contribution: HZD-Based Co-Design}

This paper presents an approach to co-design, as illustrated in Fig. \ref{fig:humanoid_front_page}, leveraging the framework of hybrid zero dynamics (HZD) and is demonstrated on a novel humanoid robot, ADAM, whose legs were designed with this process.  

Core to our approach to co-design is to utilize a gait generation method that generates walking gaits leveraging the full-order dynamics of the robot: HZD \cite{grizzle2014models}.  HZD is both rooted in formal guarantees that build upon methods from nonlinear control \cite{westervelt2003hybrid,ames2014human}, and has proven effective in generating dynamic walking gaits on humanoid robots in practice \cite{hereid20163d,buss2014preliminary,hubicki2016atrias,ambrose2017toward,apgar2018fast,tucker2021preference}.  
Key to the HZD approach is the use of \emph{virtual constraints}---these are constraints enforced via actuators that drive actual outputs to desired outputs.  The parameters of the desired outputs are then optimized to minimize a cost subject to hybrid invariance conditions on the zero dynamics, resulting in the HZD optimization problem which outputs stable gaits. 

The key idea of this paper is to extend the virtual constraint framework to include design parameters. 
In particular, we introduce \emph{design} virtual constraints where the actual joint lengths are driven to desired values determined by an HZD optimization problem. Just as virtual constraints are ``fictitous'' constraints enforced by actuators, design virtual constraints are ``fictitous'' constraints enforced by the mechanical design of the robot.  Therefore, the HZD co-design optimization problem presented (see Fig. \ref{fig:humanoid_front_page}) simultaneously designs the robot, its controllers, and the walking gait that minimizes a given cost.  This is demonstrated through the co-design of a novel humanoid robot: ADAM.  Specifically, the lengths of its limbs are co-designed via the HZD optimization problem.  The resulting walking gait is realized experimentally on the robot in the context of planar walking \cite{video}. Therefore, this paper demonstrates the end-to-end HZD co-design process. 


\section{HZD GAIT GENERATION}

A robot performing locomotion goes through multiple continuous phases separated by discrete impacts. This combination of continuous-time and discrete-time dynamics lends itself to being modeled as a hybrid system. To control such systems, the Hybrid Zero Dynamics (HZD) method \cite{grizzle2014models} has proven itself to be a powerful tool. In this section, we present the HZD framework and how it can be used together with trajectory optimization to produce stable walking trajectories for the closed-loop system. 

\subsection{Hybrid Zero Dynamics Framework}

Let the configuration coordinates of some robotic system be given by $q \in \mathcal{Q} \subset \mathbb{R}^n$. Furthermore, let the full system state be given by $x = (q, \dot{q}) \in \mathcal{X} \subset  \mathsf{T}\mathcal{Q}$. This leads to a second-order mechanical system which can be expressed in standard form as:
\begin{equation}
    D(q) \ddot{q} + H(q, \dot{q}) = Bu,
    \label{eq:EL}
\end{equation}
where $D(q) \in \mathbb{R}^{n \times n}$ is the inertia matrix, $H(q, \dot{q}) \in \mathbb{R}^n$ is the drift vector, $B \in \mathbb{R}^{n \times m}$ is the actuation matrix, and $u \in \mathcal{U} \in \mathbb{R}^m$ is the system input. In state space form the dynamical system becomes:
\begin{equation}
    \dot{x} = f(x) + g(x)u.
\end{equation}


In general, a hybrid dynamic control system can be defined as a tuple containing sets of continuous-time dynamics (referred to as domains) and discrete-time impact maps connecting the domains. While some modes of locomotion require multiple domains to model changing ground contacts \cite{reher2020algorithmic}, point-foot or flat-foot symmetric walking can be described by a symmetric single-domain hybrid system, where the domain is defined by whichever foot is the stance foot and which is the swing foot. In this paper, we deal with a point-foot humanoid robot. For clarity, we choose to describe the single-domain case here. The more general approach for multi-domain systems is outlined in \cite{reher2021dynamic}. 

Let the swing foot height above the ground be defined as $z_{nsf} : \mathcal{Q} \rightarrow \mathbb{R}$. We then have that the set of admissible states within the domain are given by:
\begin{equation}
    \mathcal{D} := \{ (q, \dot{q}) \in \mathcal{X} \ | \ z_{nsf}(q) \geq 0 \} \ \subset \mathcal{X}.
\end{equation}
The impact of the swing foot with the ground denotes the edge which triggers the transition from one domain to another. The region in which this occurs is denoted as the switching surface $\mathcal{S} \subset \mathcal{D}$ which is defined as:
\begin{equation}
    \mathcal{S} := \{(q, \dot{q}) \in \mathcal{X} \ | \ z_{nsf}(q) = 0, \ \dot{z}_{nsf}(q, \dot{q}) < 0\}.
\end{equation}
The impact of the swing foot with the ground results in a discrete change in the states which can be captured using a reset map $\Delta : \mathcal{S} \rightarrow \mathcal{X}$, which can be defined as:

\begin{equation}
    x^+ = \Delta(x^-), \quad x^- \in \mathcal{S}
    \label{eq:reset_map}
\end{equation}
where $x^-$ and $x^+$ refer to the pre- and post-impact states respectively. 

By combining the continuous dynamics within a domain and the reset map from \eqref{eq:reset_map}, the single-domain hybrid control system is obtained:
\begin{equation}
    \mathcal{HC} = 
    \begin{cases}
        \dot{x} = f(x) + g(x)u \quad & x\notin \mathcal{S} \\
        x^+ = \Delta(x^-) & x^- \in \mathcal{S}
    \end{cases}
    \label{eq:hzd_single_domain}
\end{equation}

The idea behind the HZD framework is to reduce the hybrid dynamical system in \eqref{eq:hzd_single_domain} into a lower dimensional system. Let the set of outputs that we want to control be given by $y^a(q)$, and let the desired outputs be given by $y^d(\tau(q), \alpha)$. Furthermore, we can shape the behavior of the outputs through the construction of virtual constraints $y_{\alpha}: \mathcal{Q} \rightarrow \mathbb{R}^m$:
\begin{equation}
    y_{\alpha}(q) := y^a(q) - y^d(\tau (q), \alpha).
    \label{eq:output_constraint}
\end{equation}
When the outputs and their derivatives are equal to zero, we say that the residual dynamics evolve on the zero dynamics surface, defined as:
\begin{equation}
    \mathcal{Z}_{\alpha} := \{ x \in \mathcal{D} \ | \ y_{\alpha}(q) = 0, \ \dot{y}_{\alpha}(q) = 0 \}.
    \label{eq:zero_dynamics_surface}
\end{equation}
In \eqref{eq:output_constraint}, we choose to describe the desired outputs $y^d$ using B\'ezier polynomials of order $b \in \mathbb{N}_{\geq 0}$. Explicitly, this allows us to parameterize $y^d$ using a phasing variable $\tau: \mathcal{Q} \to \mathbb{R}$ and to shape the desired outputs using a collection of Bézier coefficients $\alpha \in \mathbb{R}^{m \times (b+1)}$. 

Using a feedback controller $u^*(x)$ we can drive the outputs to zero exponentially, resulting in the closed-loop system $\dot{x} = f_{cl}(x) := f(x) + g(x)u^*(x)$. With the right choice of controller (feedback linearizing controllers are often used as discussed in Sec. \ref{sec:co_design_framework}), the system exponentially converges to the zero dynamics surface on the continuous dynamics. 

Furthermore, to ensure stability of the hybrid system as a whole, we must ensure that the outputs remain zero through impact (i.e. the outputs are impact-invariant). 
This impact-invariance condition is often termed the \textit{HZD condition}:
\begin{equation}
    \Delta (\mathcal{S} \cap \mathcal{Z}_\alpha) \in \mathcal{Z}_\alpha.
    \label{eq:HZDcond}
\end{equation}

\subsection{Trajectory Optimization}

After establishing the dynamics and control objectives of a hybrid control system, the next step is to synthesize the desired behavior of our system through the selection of coefficients $\alpha$. This can be achieved using trajectory optimization. The specific tool used in this paper is FROST \cite{hereid2017frost}, which utilizes direct collocation and has been demonstrated to successfully generate periodic gaits for various types of legged systems, including walking \cite{reher2019dynamic} and running \cite{ma2017bipedal} bipeds, quadrupeds \cite{ma2020coupled}, and exoskeletons \cite{harib2018feedback}. 

Direct collocation allows us to numerically approximate a solution to the dynamical system presented in \eqref{eq:hzd_single_domain}, and it also makes it easy to add additional constraints that need to be satisfied throughout the gait. The optimization problem to find stable periodic gaits for a legged system can be stated as follows:
\begin{align}
       \underset{\alpha, X}{\textrm{argmin}} & \ \Phi(X)
    \label{eq:standard_optimization_problem} \\
        \textrm{s.t.} \quad & \dot{x} = f_{cl}(x) \tag{Closed-Loop Dynamics} \\
        & \Delta(\mathcal{S} \cap \mathcal{Z}_\alpha) \subset \mathcal{Z}_\alpha \tag{Impact-Invariance}\\
        & X_{\min} \leq X \leq X_{\max}  \notag \tag{Decision Variables}\\
        & c_{\min} \leq c(X) \leq c_{\max} \tag{Physical Constraints}\\
        & a_{\min} \leq a(X) \leq a_{\max}\tag{Essential Constraints} 
\end{align}

In \eqref{eq:standard_optimization_problem}, $\Phi(X)$ is the cost function that specifies the term we aim to minimize, e.g. joint torques, energy efficiency, linear velocity, or a combination of multiple factors. The physical constraints ensure that the obtained gaits are physically achievable for the robot. These include friction cone constraints which prevent the robot from slipping on the ground \cite{underactuated}, workspace constraints which ensure collision-free movement, and actuator limits which guarantee that the required actuation efforts are within the actuators' bounds. The essential constraints \cite{tucker2021preference} are constraints that can be used to shape the gait behavior and in so doing restrict the search space for potential gaits. These constraints can include step length, step duration, average walking speed, minimum foot clearance, etc. When generating gaits, tuning these parameters can play a vital part in achieving robust, smooth gaits. 

\section{Co-Design Framework}\label{sec:co_design_framework}

While the optimization problem presented in \eqref{eq:standard_optimization_problem} can be used to generate stable periodic gaits for a walking robot, the framework only optimizes over the outputs (often selected to be the joints of the robot). In this section, we will introduce an approach to extend the optimization problem to also be over the space of design parameters. This approach first requires an extension of a robot model, followed by an extension of the HZD optimization problem. Additional implementation details are provided in Sec. \ref{sec:model_generation}. The resulting methodology is later demonstrated towards the design of the humanoid ADAM in Sec. \ref{sec:adam_co-design}.


\subsection{Encoding Design Parameters as Virtual Constraints}
\begin{figure}%
    \centering
    \subfloat[\centering Original\label{fig:holonomic_original}]{{\includegraphics[width=3cm]{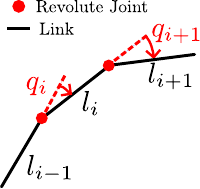} }}%
    \qquad
    \subfloat[\centering Extended\label{fig:holonomic_extended}]{{\includegraphics[width=3cm]{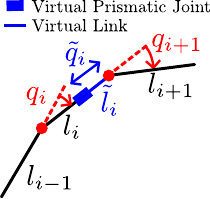} }}%
    \vspace{-0.3em}
    \caption{Visualization of how a link can be extended by adding a virtual joint and link.}%
    \label{fig:holonomic_contraints}%
    \vspace{-1.5em}
\end{figure}

Consider a system with $d \in \mathbb{N}_{> 0}$ design parameters. Specifically, these design parameters represent the lengths associated with specific robotic links. The link lengths, denoted $l_i$ for link $i \in \mathbb{N}$, can be represented as configuration coordinates by augmenting the system with $d$ prismatic \textit{virtual joints}, $\tilde{q}_i$; for an illustration of this, see Fig. \ref{fig:holonomic_original}. We term the joint $\tilde{q}_i$ a virtual joint because in the optimization problem, it is treated as any other joint, but in the real world, it has to be fixed. The virtual link $\tilde{l}_i$ is massless and serves as the parent link to the next link, $l_{i+1}$.

 The design parameters $l_i$, for $i = 1,\dots,d$, can thus be represented through the corresponding set of configuration coordinates $q_{D} \in \mathcal{Q}_D \subset \R^d$. The augmented system configuration including the design coordinates is then denoted as $\tilde{q} := (q^{\top}, q_D^{\top})^{\top} \in \tilde{\mathcal{Q}} \subset \R^{n + d}$, with the full augmented system state $\tilde{x} = (q^{\top}, q_D^{\top},\dot{q}^{\top},\dot{q}_D^{\top}) \in \tilde{\mathcal{X}} \subset \mathsf{T}\tilde{\mathcal{{Q}}}$. It is important to note that this augmented system has corresponding augmented continuous-time dynamics which can be expressed using the Euler-Lagrange equations applied to the augmented configuration coordinates $\tilde{q}$. We will refer to the resulting augmented dynamics as $\tilde{x} = \tilde{f}(\tilde{x}) + \tilde{g}(\tilde{x})u$.

Following this notation, the design parameters can be shaped along with the nominal outputs by augmenting the virtual constraints to be the following:
\begin{align}
    y(\tilde{q}) := \begin{bmatrix}
        y_{\alpha}(q) \\ y_{\beta}(q_{D})
    \end{bmatrix} = \begin{bmatrix} y^a(q) - y^d(\tau(q),\alpha) \\ y^a_{D}(q_D) - \beta \end{bmatrix},
\end{align}
where $y_{\alpha}(q)$ are the nominal virtual constraints as defined in \eqref{eq:output_constraint}, and $y_{\beta}(q_D)$ are a new set of virtual constraints  termed \textit{design virtual constraints}, $y_{\beta}: {Q}_D \to \R^d$. As before, $y^a_D(q_D)$ denotes the actual design outputs, and $\beta \in \R^d$ denotes the desired design outputs. 

As with the nominal trajectory optimization procedure, these virtual constraints are driven to zero exponentially through the use of an Input/Output Linearization (also called a Feedback Linearizing) controller. Explicitly, in the case where $y_{\alpha}(q)$ consists of all relative degree 2 (position-modulating) outputs, the form of this controller is:
\begin{align}
    u_{D}(\tilde{x}) := -A^{-1}(\tilde{x}) \left( L^2_{\tilde{f}}y(\tilde{x}) + 2\varepsilon L_{\tilde{f}}y(\tilde{x}) + \varepsilon^2 y(\tilde{q}) \right),
\end{align}
with control gain $\varepsilon > 0$ and $A(\tilde{x}) := L_{\tilde{g}}L_{\tilde{f}}y(\tilde{x})$. Here, $L$ denotes the Lie derivative. Explicitly, since our outputs are relative degree 2, our Lie derivatives are expressed as $L_fy(x) = \frac{\partial y(q)}{\partial q}\dot{q}$, $L^2_fy(x) = \frac{\partial}{\partial q} \left (\frac{\partial y(q)}{\partial q}\dot{q} \right) \dot{q}$, and $L_gL_fy(x) = \frac{\partial y(q)}{\partial q}\ddot{q}$.

\subsection{Constraining the Virtual Joints}\label{subsec:virtual_joints}
As with the physical joints of the robot, the virtual joints $\tilde{q}_i$ have some physical limits. Here, their bounds depend on the minimum and maximum allowable length of the corresponding limb length $l_i$. These limits are enforced along with the other physical constraints in the form as
\begin{equation}
    q_{D,\min} \leq q_D \leq q_{D, \max}.
\end{equation}
Note that later we will lump these constraints in with the other physical constraints of the system. Furthermore, it is important to note that since the design virtual joints $q_D$ will be fixed on the actual robot it may not change throughout the gait, meaning that the velocity of the virtual joint must be kept equal to zero, i.e. $\dot{q}_D = \bm{0}_{d}$.

\subsection{Co-Design Optimization Problem}
Lastly, we will explicitly state the co-design optimization problem in a way that can be numerically approached (as done in \cite{ames2014human}). First, we can denote the virtual constraints evaluated when the generated orbit intersects the guard as:
\begin{align}
    v(\alpha,\beta) = \tilde{q} \quad \textrm{s.t.}\quad  \begin{bmatrix}
        y(\Delta_{\tilde{q}} \tilde{q}) \\ z_{nsf}(\tilde{q})
    \end{bmatrix} = \begin{bmatrix}
        \bm{0}_{m+d} \\ 0
    \end{bmatrix},
\end{align}
where $\Delta_{\tilde{q}}$ is the relabeling matrix corresponding to the augmented configuration coordinates $\tilde{q}$. Since typically humanoids have symmetric limb lengths, then this relabeling matrix can be generally expressed as:
\begin{align}
    \Delta_{\tilde{q}} := \begin{bmatrix}
        \Delta_q & \bm{0}_{m \times d} \\
        \bm{0}_{d \times m} & I_{d \times d}
    \end{bmatrix}.
\end{align}

We will similarly denote the first-order derivative of this intersected point as:
\begin{align}
    \dot{v}(\alpha,\beta) = \{ \tilde{x} ~\textrm{s.t.}~
        &L_{\tilde{f}}y(\tilde{\Delta}(\tilde{x})) = \bm{0}_{m+d}, \dot{z}_{nsf}(\tilde{x}) < 0 \},
\end{align}
where $\Tilde\Delta$ is the impact map associated with the augmented system.

Using this notation, the impact-invariance condition \eqref{eq:HZDcond} can be replaced by the following conditions:
\begin{align}
    y(v(\alpha,\beta)) = \bm{0}_{m+d}, \\
    L_{\tilde{f}}y(\dot{v}(\alpha,\beta)) = \bm{0}_{m+d}.
\end{align}

Adding the augmented constraints to the standard HZD optimization problem presented in \eqref{eq:standard_optimization_problem} leads to the following extended optimization problem:
\begin{align}
         \underset{\alpha, \beta, X}{\textrm{argmin}} & \ \Phi(X)
        \label{eq:extended_optimization_problem} \\
        \textrm{s.t.} \quad & \dot{\tilde{x}} = \tilde{f}(\tilde{x})+\tilde{g}(\tilde{x})u_D(\tilde{x}) \tag{Closed-Loop Dynamics} \\
        & y(v(\alpha,\beta)) = \bm{0}_{m+d}, \notag \\
        & L_{\tilde{f}}y(\dot{v}(\alpha,\beta)) = \bm{0}_{m+d}  \tag{Impact-Invariance} \\
        & X_{\min} \leq X \leq X_{\max}  \notag \tag{Decision Variables}\\
        & c_{\min} \leq c(X) \leq c_{\max} \tag{Physical Constraints}\\
        & a_{\min} \leq a(X) \leq a_{\max} \tag{Essential Constraints}
\end{align}

\subsection{Accounting for Changing Inertia}\label{subsec:changing_mass}
The extended optimization problem in \eqref{eq:extended_optimization_problem} can be used to optimize over link lengths during the gait generation, however, in its current form it does not account for changing inertial properties in the augmented links as the link lengths change. Because the virtual links $\tilde{l}_i$ are massless, the inertial properties of the augmented links are completely determined by $\alpha$. The optimization problem is therefore solved by assuming fixed inertial properties for the augmented links. To overcome this problem, an iterative outer loop can be used during the optimization procedure. 

\begin{algorithm}[b]
\caption{Varying Inertial Properties}\label{alg:inertia}
\begin{algorithmic}
\State {$q_{D,curr} = q_{D,inital}$}
\While{$|| q_{D,curr} - q_{D,prev} || > \delta$} 
    \State {$\texttt{model} = \texttt{GenerateModel}(q_{D,curr})$}
    \State {$q_{D,prev} = q_{D,curr}$}
    \State {$q_{D,curr} = \texttt{SolveOptimizationProblem(model)}$}
\EndWhile
\State {$q_{D,final} = q_{D,curr}$}
\end{algorithmic}
\end{algorithm}

The iterative algorithm is presented in Alg. \ref{alg:inertia}. In short, the algorithm updates the inertial properties of the robot model based on the latest optimized virtual joint values. This ensures that the robot model being optimized has precise inertial values. The iterative loop is exited when convergence on the virtual joints is reached, i.e. once the change in virtual joints is smaller than some threshold, $\delta$. The addition of this algorithm is particularly important if the inertial properties change drastically with the link lengths, such as if the virtual joint space is large.

\section{Co-Design of the Humanoid Robot ADAM}\label{sec:adam_co-design}
In this section, we demonstrate how the HZD Co-Design framework was used to decide the optimal leg lengths of the 20 degrees of freedom (DoFs) humanoid robot ADAM. Specifically, we wanted to find the optimal thigh and shin lengths for the robot, when performing planar locomotion. To illustrate the process, we first present the robot model and how it is augmented. Following this, we present some practical implementation details on how the optimization problem was solved. Finally, we present the results of the method.

\subsection{Augmenting the Robot Model}

\begin{figure}%
    \centering
    \subfloat[\centering Original\label{fig:humanoid_dofs_original}]
    {{\includegraphics[width=0.3\linewidth]{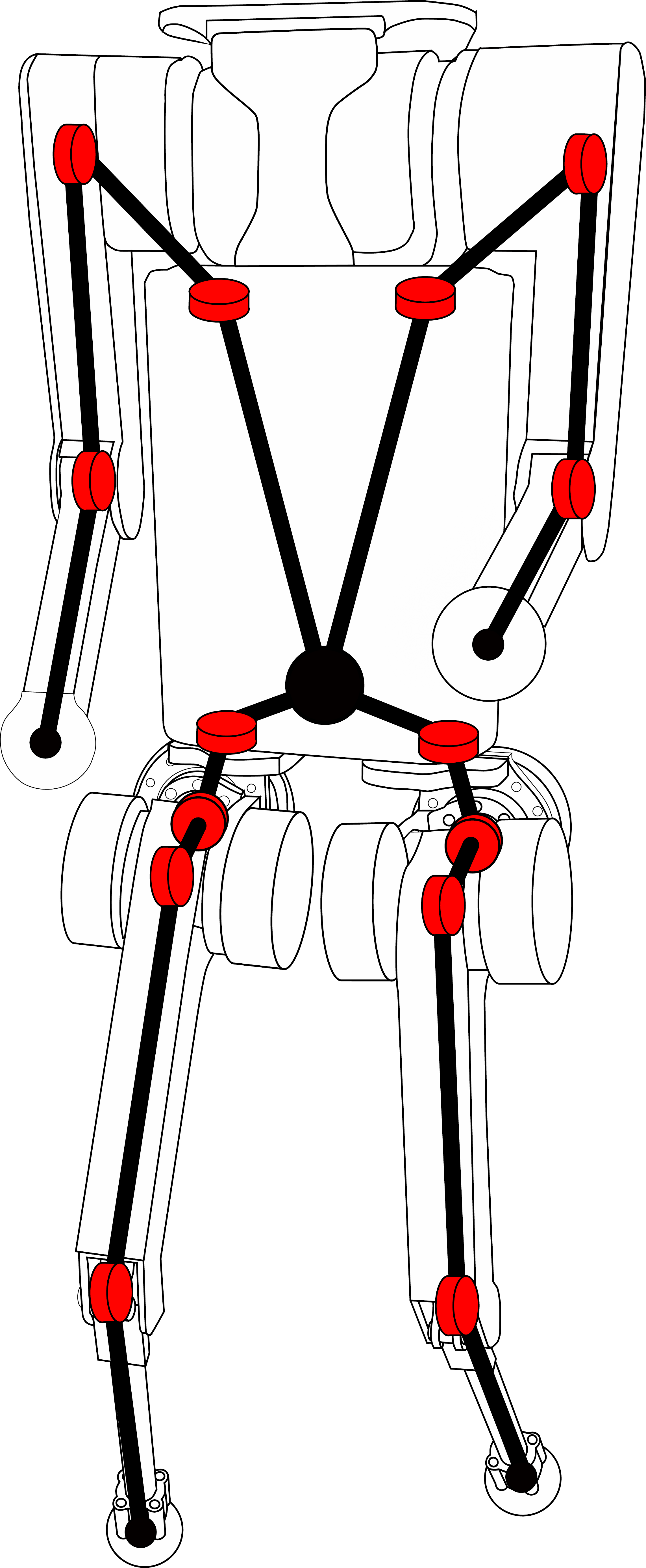} }}%
    \qquad
    \subfloat[\centering Extended\label{fig:humanoid_dofs_extended}]
    {{\includegraphics[width=0.3\linewidth]{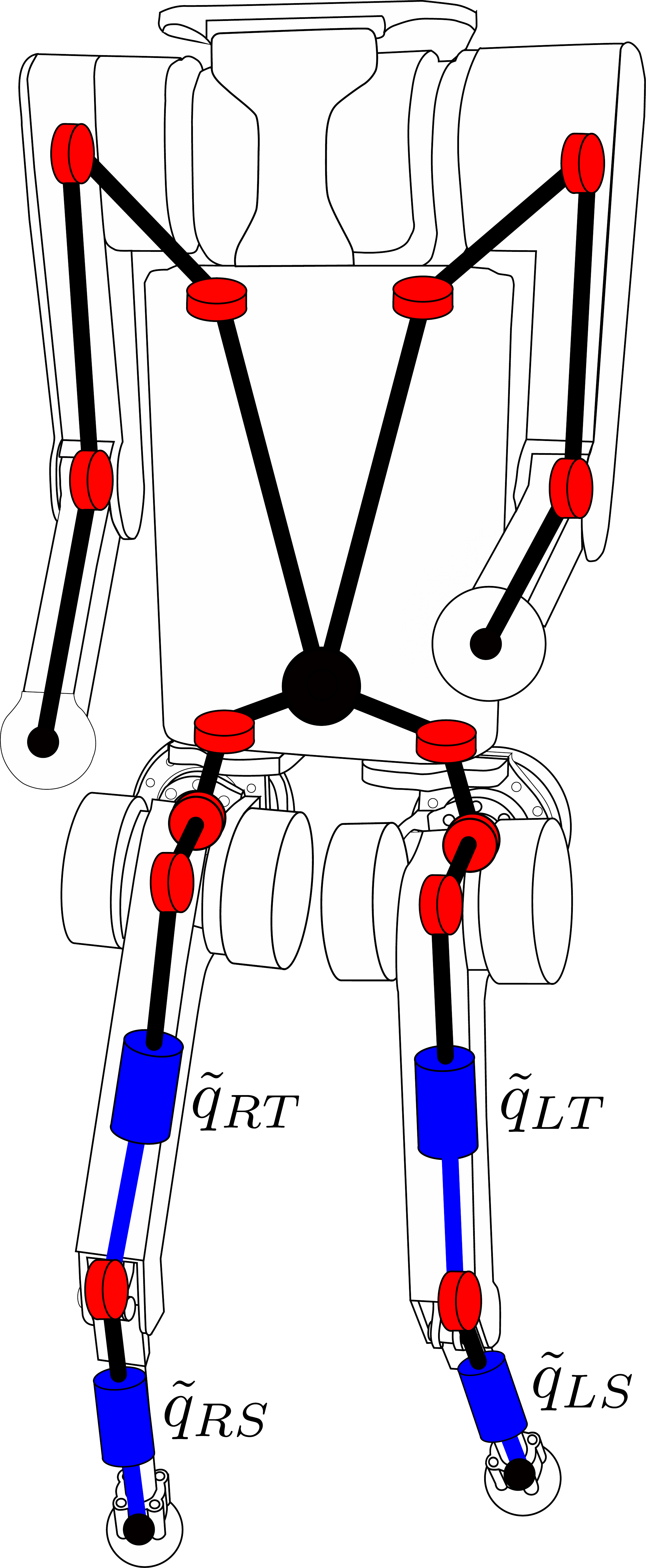} }}%
    \caption{The ADAM humanoid's original and augmented models after adding virtual joints to the thighs and shins.}%
    \label{fig:humanoid_dofs}%
    \vspace{-1.5em}
\end{figure}

ADAM is 20 DoFs humanoid robot that was developed by taking the torso and actuators from an A1 Unitree quadrupedal robot, and then by designing the legs and arms from scratch. Each of the legs have 4 actuated DoFs, where the innermost actuator controls the yaw, the next controls the roll, and then the last two control the hip pitch and knee pitch joint respectively. Each of the arms has 3 actuated DoFs, where the innermost actuator controls the yaw of the arm, the second actuator controls the pitch of the upper arm, and the last actuator controls the pitch of the forearm. Fig. \ref{fig:humanoid_dofs_original} shows an overview of the robot model.

Because we wish to optimize over the thigh lengths and shin lengths, it is necessary to add four virtual joints to the original robot model, i.e. $q_D = [q_{LT}, q_{RT}, q_{LS},q_{RS}]^{\top}$ as illustrated in Fig. \ref{fig:humanoid_dofs_extended}. This model augmentation results in a 24 DoFs robot model. Furthermore, because we want the robot to have equal leg lengths, we also enforce that the left and right links should be equal. That is, the following constraints are added to the physical constraints of the optimization problem: $q_{LT}= q_{RT}$, $q_{LS} = q_{RS}$.

\subsection{Automatic Robot Model Generation}\label{sec:model_generation}
To solve the augmented optimization problem in \eqref{eq:extended_optimization_problem} using FROST, it is necessary to provide the program with a Unified Robot Description Format (URDF) model, which is an XML-formatted file containing a complete description of the robot's kinematic and inertial properties. Normally, when using FROST to generate gaits, a non-changing URDF is used. That is, the URDF model is created for the robot and then used for all the following gait generation procedures. Because of the iterative process from Alg. \ref{alg:inertia}, it is necessary to be able to generate URDF models with accurate inertial data for arbitrary virtual joint values. That is, given $q_D$, we must be able to generate a robot model with kinematic and inertial properties corresponding to the virtual joint values. To achieve this a function $\texttt{GenerateModel}(q_{D})$ was created.

Since the inertial properties of the links depend on their lengths, it was necessary to obtain a function relating each of the inertial properties to the link lengths. For simple geometric shapes, this can be done exactly, but as each of the links have very complex geometry such calculations were intractable. To circumvent this we employed an empiric approach where we uniformly extracted the inertial properties for a handful of different thigh and shin lengths, and then used 3rd-order polynomials to fit the data. Despite the small number of inertial property samples, we were able to fit the polynomials with near-perfect accuracy.


\subsection{Solving the Optimization Problem}
To solve the optimization problem, the IPOPT solver \cite{wachter2006implementation} in MATLAB was used. The augmented optimization problem for the 24 DoFs robot is highly nonlinear and non-convex. Since we chose to optimize for planar walking, the joints not acting in the sagittal plane were fixed. This effectively reduced the robot model to a 15 DoFs robot. 

An important task in nonlinear optimization is to provide the solver with a good initial guess. For the humanoid robot, we choose to use a two-step approach to achieve this. The first initial guess was obtained by fixing the virtual joint values and by optimizing without a cost function. This solution was then used to obtain a second initial guess where the cost function was included, while the virtual joints remained fixed. The initial guess obtained after going through the two steps greatly aided the solver in finding solutions for the full-scale nonlinear optimization problem.

\subsection{Optimization Results}

\begin{table}
\caption{Essential Constraints used in the HZD Optimization Problem.}
\label{tab:gait_constraints}
\vspace{-1em}
\begin{center}
\begin{tabular}{|c|c|c|}
\hline
Variable & Lower Limit & Upper Limit\\
\hline
Step Length (m) & $0.15$ & $0.35$ \\
\hline
Step Duration (s) & $0.6$ & $0.9$ \\
\hline
Thigh Length (m) & $0.15$ & $0.35$ \\
\hline
Shin Length (m) & $0.15$ & $0.35$ \\
\hline
\end{tabular}
\end{center}
\vspace{-1em}
\end{table}

\begin{figure}
    \centering
    \includegraphics[width=0.97\linewidth]{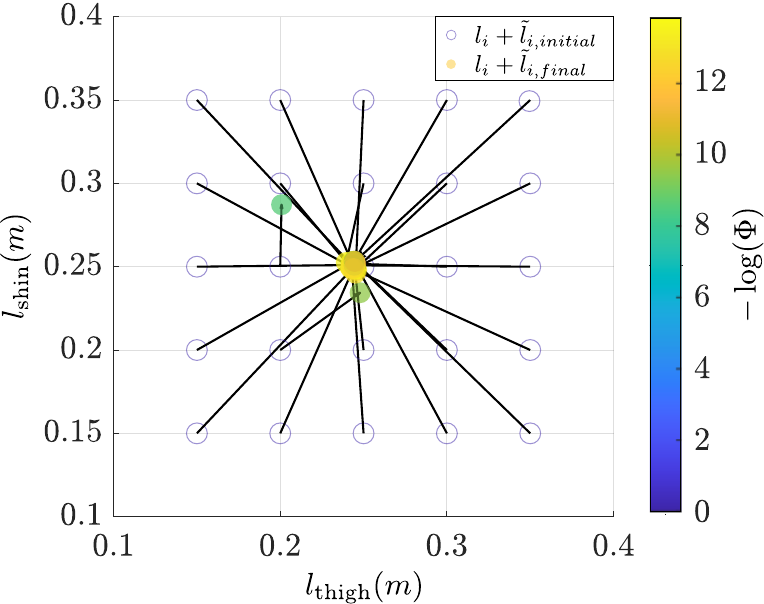}
    \vspace{-0.5em}
    \caption{A visualization of how the link lengths and thigh lengths converge, with $l_{\{\textrm{thigh},\textrm{shin}\}}$ denoting the total link lengths (i.e. $l_i + \tilde{l}_i$), from various initial values when using HZD Co-Design algorithm.}
    \label{fig:convergence_of_lengths}
    \vspace{-1.5em}
\end{figure}

The humanoid co-design optimization problem was run with the parameters listed in Table \ref{tab:gait_constraints} from a large variety of initial thigh and shin lengths, ranging from the minimum to maximum link lengths. The chosen cost metric to minimize was weighted mechanical cost of transport, in order to maximize the energy efficiency of the humanoid. Fig. \ref{fig:convergence_of_lengths} visualizes how the link lengths generally converge very closely to the same values irrespective of their initial values. In the figure the hollow circles indicate the initial values, the colored dots indicate the converged values, and the black lines relate the initial and converged values. Of the 25 runs it can be seen that in a couple of instances, the leg lengths converged to local minima. However, from analyzing the cost functions, it is clear that these two cases result in worse walking behavior with lower energy efficiency. Table \ref{tab:optimization_results} shows several of the key parameters from the optimization. From the table, it can be seen that the optimal thigh length was found to be $0.24$ (m), which is slightly shorter than the optimal shin length of $0.25$ (m). Both variables converged with high accuracy with standard deviations below $0.01$.  

\begin{table}
\caption{Statistics of the optimal gaits generated using various initial design link lengths.}
\label{tab:optimization_results}
\vspace{-1em}
\begin{center}
\begin{tabular}{|c|c|c|}
\hline
Parameter & Mean & Std \\
\hline
Thigh Length (m) & $0.2425$ & $0.0088$ \\
\hline
Shin Length (m) & $0.2515$ & $0.0084$ \\
\hline
Step Length (m) & $0.2281$ & $0.0281$ \\
\hline
Step Duration (s) & $0.7994$ & $0.0165$ \\
\hline
Walking Speed (m/s) & $0.2853$ & $0.0338$ \\
\hline
\end{tabular}
\end{center}
\vspace{-2mm}
\end{table}




\subsection{Experimental Validation}

To evaluate the co-design framework, the optimal planar gait from FROST was realized on the optimized hardware design. The optimized hardware design for ADAM features a thigh and shin length of 0.24 (m) and 0.25 (m), respectively. The humanoid was placed on a treadmill with a boom restricting the motion to the sagittal plane.
Note that the additional mass and inertia of the boom were not accounted for in the gait generation. The step length and step duration of the optimal gait as assessed in simulations and experiments are 0.23 (m) and 0.8 (s), respectively.
%

\begin{figure}%
    \centering
    \includegraphics[width=0.97\columnwidth]{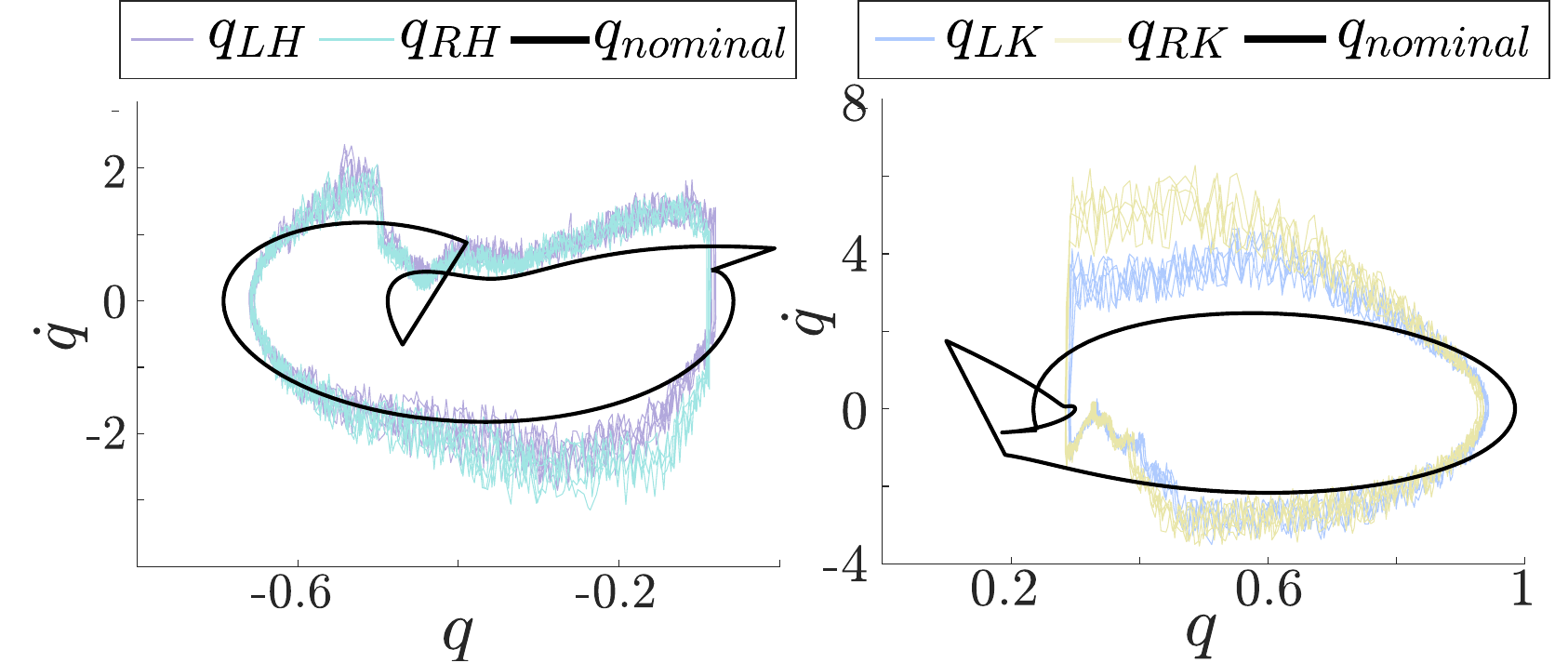}
    \vspace{-0.5em}
    \caption{Phase portraits of the simulation and hardware results with $q_{LH,RH}$ and $q_{LK,RK}$ denoting the left/right hip/knee.}%
    \label{fig:phase_plots}%
    \vspace{-2em}
\end{figure}

\begin{figure*}
    \centering
  \includegraphics[width=0.92\textwidth]{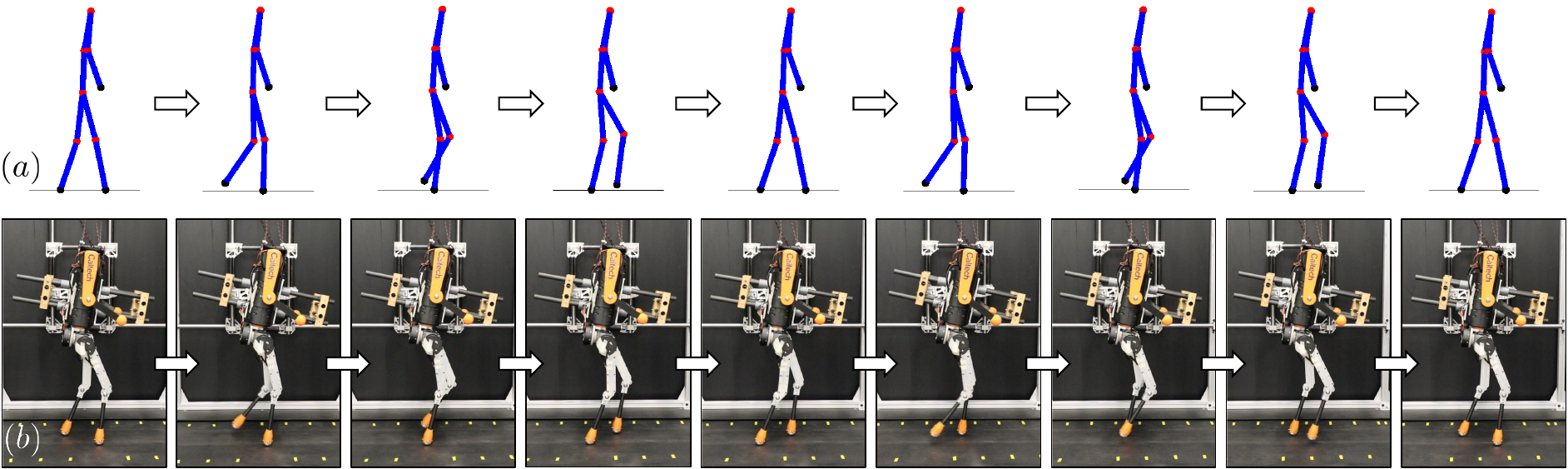}
  \caption{Gait tiles that describe the evolution of the walking throughout the gait cycle (a) in simulation and (b) on hardware.}
  \label{fig:gait_comparison}
  \vspace{-1.8em}
\end{figure*}


The generated gait was enforced on the optimized ADAM platform using joint-level PD control and using an inertial measurement unit (IMU) to estimate the global pitch of the torso. The experimental walking is visualized via output phase portraits in Fig. \ref{fig:phase_plots} and via gait tiles in Fig. \ref{fig:gait_comparison}. The experimental walking was stable and robust; the stability reflected by the robot's ability to walk for an extended period as shown in the phase portraits, and the robustness illustrated by the humanoid being able to remain stable at speeds ranging from $0.2$ to $0.3$ (m/s). The similarities between the simulated gait and the hardware gait can also be seen in Fig. \ref{fig:gait_comparison}, where the motion of the gait at various stages is compared. A video showcasing the experimental results is provided online \cite{video}.





\section{CONCLUSION} 

In this paper, we present a novel co-design framework for legged robots by coupling methods from nonlinear control theory and trajectory optimization. We demonstrate the co-design framework towards the design of the 20 DoFs humanoid ADAM and experimentally deploy the optimized gaits to realize stable and robust planar walking. The proposed model augmentation and its incorporation directly into the gait optimization problem enables simultaneous optimization of robot design parameters with nominal periodic orbits. Compared to existing co-design frameworks, our approach also leverages the full model of the robot which eliminates the need to make simplified model assumptions. However, a limitation associated with using the full-order model is that the resulting optimization problem is highly nonlinear and can be influenced by the choice of essential constraints. Future work includes extending the planar walking to 3D walking and evaluating the co-design framework across other legged platforms.




\bibliographystyle{IEEEtran}
\bibliography{99_references}

\end{document}